\documentclass[a4paper,twoside]{article}
\usepackage{etex}
\usepackage[T1]{fontenc}
\usepackage[latin1]{inputenc}
\usepackage[english]{babel}
\usepackage{graphicx}
\usepackage[noindentafter,pagestyles]{titlesec}
\usepackage{textcomp}
\usepackage{dcolumn}
\usepackage{amsmath}
\usepackage{txfonts}
\usepackage{fancyhdr}
\usepackage{threeparttable}
\usepackage{xspace}
\usepackage{cite}
\usepackage{subfigure}
\usepackage{pictex}
\usepackage{svg}
\usepackage{xfrac}
\usepackage{booktabs}
\usepackage{siunitx}
\svgpath{Images/}
\newcommand{\bmp}[1]{\begin{minipage}{#1\columnwidth}}
\newcommand{\emp}{\end{minipage}}

\newcommand{\bea}{\begin{eqnarray}}
\newcommand{\eea}{\end{eqnarray}}
\newcommand{\be}{\begin{equation}}
\newcommand{\ee}{\end{equation}}

\titleformat{\section}{\large\bfseries}{\thesection.}{.3em}{}
\titlespacing*{\section}{\leftmargini}{*3}{*3}
\titleformat{\subsection}{\bfseries}{\thesubsection}{.3em}{}
\titlespacing*{\subsection}{0pt}{*3}{*3}
\makeatletter
\def\@maketitle{%
  \newpage
  \null
  \vskip 2em%
  \begin{center}%
  \let \footnote \thanks
    {\fontsize{18}{22}\fontseries{b}\selectfont \@title \par}%
    \vskip 1.5em%
    {\normalsize
      \lineskip .5em%
      \begin{tabular}[t]{c}%
\@author
      \end{tabular}\par}%
    \vskip 1em%
    {\large \@date}%
  \end{center}%
  \par
  \vskip 1.5em}
\renewenvironment{abstract}{%
\if@twocolumn
\section*{\abstractname}%
\else
\quotation
\noindent{\bfseries\large \abstractname\vspace*{.3ex}\par}
\fi}
{\if@twocolumn\else\endquotation\fi}
\makeatother

\fancypagestyle{plain}{%
\fancyhf{} 
\fancyhead[L]{\small \MakeUppercase{\hspace*{15mm} 11$^{\mbox{\tiny th}}$ European Conference for Aeronautics and AeroSpace Sciences (EUCASS)} \\ \quad \\ DOI: 10.13009/EUCASS2025-285}
\fancyfoot[L]{\small Copyright \copyright\ 2025 by M\"uller et al.. Published by the EUCASS association with permission.}%
} 
\begin{document}
%
%
%
%
%
\title{Reducing Experimental Testing in Space Propulsion Film Cooling Analyses by Pixelwise Generative Image Interpolation}
\author{\itshape{Adam T. M\"uller$^{\star\dag}$}, Philipp J. Teuffel$^\ddagger$, Konstantin Manassis$^\ddagger$ and Nicolaj C. Stache$^\star$ \\
\itshape$^\star$Heilbronn University of Applied Sciences, Center for Machine Learning,\\\itshape Max-Planck-Str. 39, 74081 Heilbronn, Germany \\
\itshape$^\ddagger$German Aerospace Center (DLR), Institute of Space Propulsion\\
\{adam-theo.mueller, nicolaj.stache\} @hs-heilbronn.de $\cdot$  \{philipp.teuffel, konstantin.manassis\} @dlr.de\\
$^\dag$Corresponding author}

\date{}
\maketitle
\begin{abstract}
\noindent
We propose a machine learning approach for image regression from sparse experimental measurements. We show the application of the proposed method on film cooling studies in propulsion system development, aiming to reduce the need for extensive physical testing. Our method employs a lightweight feed-forward neural network with positional encoding to generate images conditioned by input parameters. Validated on real and synthetic data, it achieves high image similarity (RMSE < 8 \%, SSIM > 93 \%) while maintaining accuracy with a 30 \% reduction of measurements. We further propose a knowledge-informed extension for local adaptability of the generated images. This approach significantly reduces required tests while preserving high-quality data, enabling efficient optimization of coolant injector configurations with applications beyond aerospace.
\end{abstract}

\section{Introduction}
Film cooling remains a vital thermal management strategy in aerospace propulsion systems, where extreme thermal loads can compromise structural integrity and operational longevity. By injecting a protective layer of coolant along hot surfaces, such as combustion chamber walls and nozzles, film cooling mitigates material stress and extends component life~\cite{shine_review_2018}. The performance of these systems, however, is governed by a complex interplay of design and operating parameters, including injector geometry, coolant properties, and local flow conditions~\cite{tauchi_optimal_2024}. Capturing and understanding these interactions experimentally requires extensive testing, which is often limited by time, cost, and resource constraints--particularly when scaling to high-dimensional parameter spaces.

To manage this complexity, surrogate modeling and data-driven approaches have increasingly complemented traditional experimental and numerical methods in aerospace engineering. Among these, machine learning has emerged as a powerful tool for reducing experimental workloads by enabling predictive modeling, parameter inference, and high-fidelity interpolation between sparse measurements. In particular, tasks involving image-based data, such as flow visualizations, schlieren photography, or surface temperature maps, offer an opportunity to apply deep learning models to capture spatially distributed phenomena in a data-efficient manner~\cite{brunton_data-driven_2021}.

This work explores the application of machine learning to reduce the experimental testing load in film cooling studies for propulsion systems. Specifically, we propose a method for conditional image interpolation, enabling the generation of synthetic coolant film patterns based on a limited set of real-world measurements. Our method utilizes a feed-forward neural network that accepts physical operating parameters and pixel-wise positional information as inputs, generating images that interpolate those acquired through experimental testing.

To demonstrate its effectiveness, the method is applied to real-world measurements of cold flow experiments on film cooling--a system of interest in ongoing green propellant development efforts at the German Aerospace Center (DLR)~\cite{kirchberger_green_nodate, teuffel_HTP_2024}. The model is trained on both synthetic and real experimental datasets and evaluated using quantitative metrics such as root mean square error (RMSE) and structural similarity index (SSIM).

Beyond pure regression, we introduce a knowledge-informed extension that enables local refinement of the generated images. Using tools such as gradient-weighted class activation mapping (Grad-CAM)~\cite{selvaraju_grad-cam_2017}, domain experts can guide and improve model performance in regions of interest, allowing for targeted corrections and enhanced interpretability.

By combining data-driven regression with domain-specific knowledge, this approach presents a scalable and efficient solution for experimental campaigns where image-based diagnostics are a core component. It enables more agile exploration of the design space, reduces reliance on costly test campaigns, and sets the stage for future applications in high-fidelity modeling across propulsion and aerospace research.
\\
\\
The main contributions of our work are summarized as follows:
\begin{itemize}
\item A general yet lightweight approach for pixel-wise image interpolation based on positional encoding, enabling accurate interpolation from sparse experimental measurements.
\item A knowledge-informed extension that enables local refinement of generated images based on expert domain knowledge.
\end{itemize}

\section{Related Work}

\textbf{Film Cooling Studies and Abstractions.}
Film cooling remains a crucial technology for thermal protection in liquid rocket engines and other high-enthalpy propulsion systems. However, experimental investigations are often tailored to specific engine configurations and propellant pairings, limiting the generalizability of their findings~\cite{zhang_thermal_2023, tauchi_optimal_2024}. While public datasets are rare, efforts do exist that aim to abstract and simplify the complex physical behavior for broader applicability~\cite{shine_review_2018}.

Several studies use analytical or surrogate modeling approaches to approximate film cooling performance in order to reduce the experimental burden~\cite{tauchi_optimal_2024, inoue_evaporation_2022}. While these methods vary in complexity, many rely on experimental configurations that isolate specific aspects of the cooling process to enable controlled, repeatable observations and data acquisition. A noteworthy example of an experimental study is presented by Kang et al.~\cite{kang_experimental_2017}, who investigate coolant film dynamics using both water and high-test peroxide (HTP) as simulants in cold flow experiments. Their setup employs a transparent surrogate combustion chamber, enabling optical access to the entire flow field--an approach similar to our use of visual diagnostics.

Beyond propulsion, insights into fluid-wall interaction can be drawn from more general studies of impinging jets. For instance, Chee et al.~\cite{chee_flow_2023} show wetting dynamics on flat surfaces exhibit characteristic zones such as the radial flow zone (RFZ), where the fluid spreads laterally from the point of impact, followed by peripheral film accumulation and drainage. These regimes are sensitive to parameters like impingement angle, impact velocity, and surface texture--mirroring the complexity encountered in film cooling applications.

\vspace{1em}
\noindent \textbf{Machine Learning in Aerospace Applications.}
In recent years, machine learning (ML) has gained traction in aerospace and space propulsion domains, particularly for its potential to reduce reliance on expensive physical testing. Common applications include anomaly detection and fault diagnosis in datasets, and surrogate modeling of complex physical systems~\cite{brunton_data-driven_2021}.

Examples include ML-driven prediction of injector behavior~\cite{zapata_usandivaras_data_2022}, data-driven modeling of hybrid propulsion systems~\cite{zavoli_surrogate_2022}, and efforts toward autonomous control in spacecraft systems~\cite{horger_experimental_2024, waxenegger-wilfing_machine_2021}. In each of these cases, ML is used to extract actionable insights or replace first-principles models when full-scale simulations or experiments are infeasible.

When working with image data, most ML applications focus on inference--extracting information from experimental images. In contrast, generating image data from input conditions is a more recent research direction. Generative approaches such as variational autoencoders (VAEs), generative adversarial networks (GANs), and diffusion models have demonstrated promising results in tasks ranging from data augmentation to realistic simulation synthesis~\cite{zhan_conditional_2025}. While powerful, these models are typically complex, require extensive training data, and often lack interpretability.

\vspace{1em}
\noindent \textbf{Generative and Interpolative Image Models.}
Generating images conditioned on parameter inputs can be framed as a regression task over high-dimensional spatial data. Generative models capable of interpolation fall broadly into two categories. Spatial interpolation models, such as PixelRNNs~\cite{oord_pixel_2016}, predict pixel values based on local context within an image. Temporal interpolation, commonly explored in video frame prediction or in-betweening tasks, often leverages GANs or diffusion-based architectures~\cite{reda_film_2022, dong_video_2023}.

Some work aims to improve generative interpolation through low-rank or adaptive latent space extensions to diffusion models~\cite{zhang_diffmorpher_2024}, enabling more realistic intermediate samples. Similarly, van den Oord et al.~\cite{van_den_oord_conditional_2016} introduced a conditional PixelCNN that allows for interpolation between known states via autoregressive pixel modeling.

Closer to our approach, Hwang et al.~\cite{hwang_machine-learning_2021} propose a lightweight pixel-wise regression model where each pixel's output is predicted independently using a condition-aware weighting vector. This enables interpolation across input parameter space without requiring extensive architectural complexity.

\begin{figure}[tb]
\centering
\subfigure[]{ 
  \makebox[0.42\textwidth]{
    \includegraphics[height=27mm]{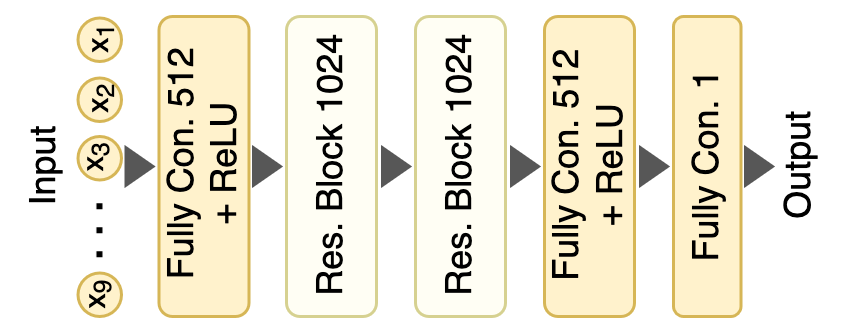}
  }
  \label{figNN:Net}
}
\hspace{3mm}
\subfigure[]{ 
  \makebox[0.52\textwidth]{
    \includegraphics[height=25mm]{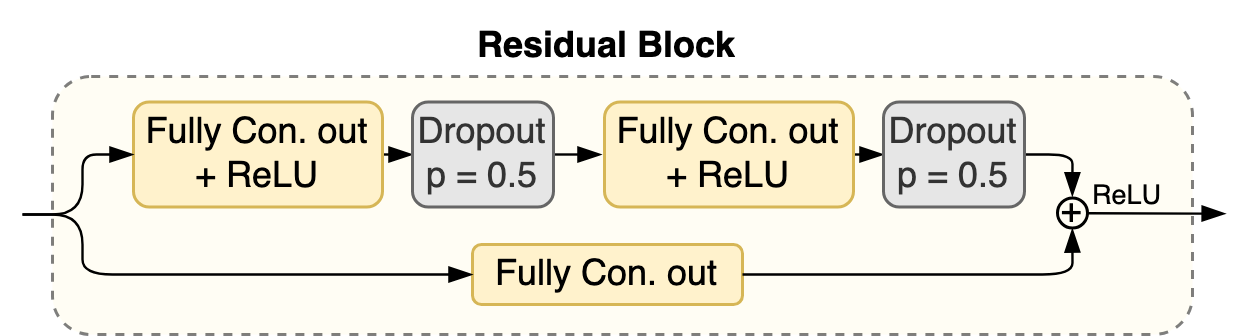}
  }
  \label{figNN:ResBlock}
}
\caption{Neural network architecture, where (a) shows the full architecture from input parameters to output and (b) depicts a residual block as implemented in (a).}
\label{figNN:main}
\end{figure}

\begin{figure}[tb]
\centering
\includegraphics[width=\textwidth]{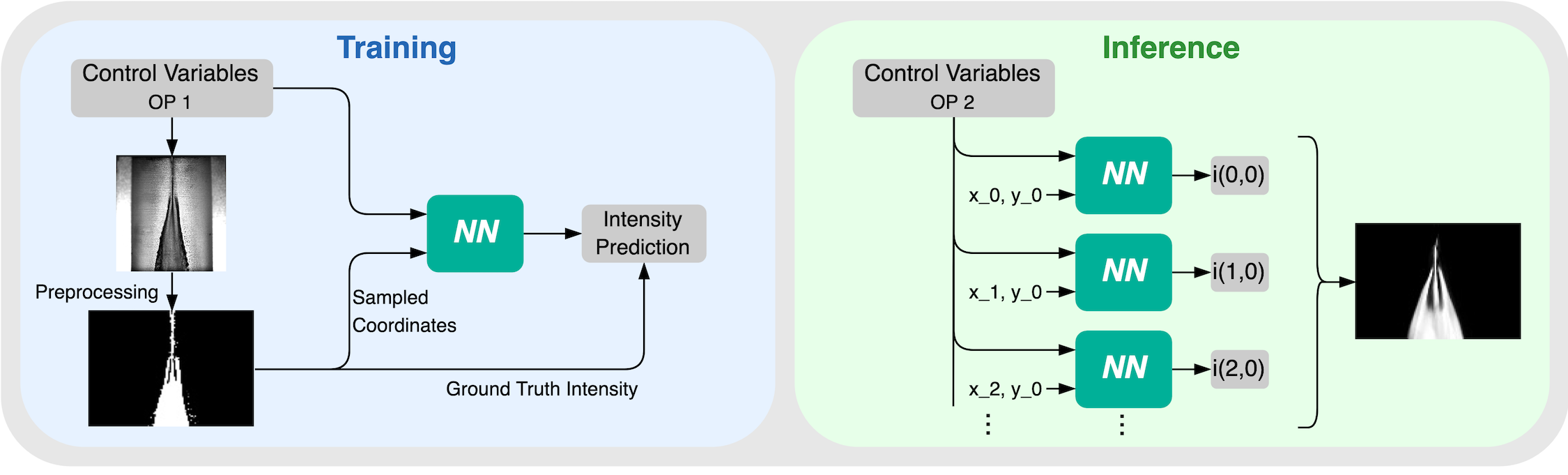}
\caption{Schematic diagram of the proposed neural network, illustrating (left) the training phase, where the network learns on sampled pixels from ground truth images, and (right) the inference phase, where the trained model generates images conditioned on input parameters of a new operating point (OP).}
\label{PixCOIN_Method}
\end{figure}

\section{Machine Learning Method} \label{sec:method}
We introduce \textbf{PixCOIN} (Pixel-based Conditional Image Interpolation), a machine learning approach for reconstructing image data from sparse physical testing data. Unlike conventional interpolation techniques or surface modeling approaches, PixCOIN operates directly in pixel space, enabling the generation of complex, spatially detailed images without relying on geometric assumptions or spatial continuity constraints.

Inspired by the conditional pixel-wise regression framework of Hwang et al.~\cite{hwang_machine-learning_2021}, our method departs in key aspects to suit the requirements of experimental propulsion environments. Most notably, we integrate normalized pixel coordinates as explicit input features, serving as an implicit form of positional encoding. This allows the model to learn spatially-dependent patterns directly, without resorting to convolutional architectures.

Conventional data reduction methods-such as scalar value approximations (e.g. contour-based or maximum film width estimations), or response surface modeling-struggle with local anomalies, discontinuities, or turbulent features in real-world film cooling behavior~\cite{gonzalez_digital_2017}. These limitations are bypassed by PixCOIN, which treats image generation as a regression task over the joint space of physical input parameters and spatial location. This enables accurate modeling of highly non-uniform cooling patterns essential for robust system optimization, without imposing assumptions of spatial continuity or regularity.

\subsection{Model Architecture and Training} \label{subsec:ArchitectureAndTraining}
PixCOIN uses a fully connected feed-forward neural network (FFNN) with ReLU activations. This architecture was enhanced by residual blocks, since results in preliminary network architecture studies showed a clear performance gain by employing a simple residual architecture. The final model consists of three dense layers with two residual blocks, comprising approximately 3.5 million trainable parameters (see Figure~\ref{figNN:main}). Model scaling experiments revealed slight performance gains when increasing from a small model ($\sim 925$ K parameters) to a medium ($\sim 3.5$ M parameters) or large model ($\sim 6.7$ M parameters). However, the difference between the medium and large models was negligible, favoring the medium model due to its more efficient use of computational resources and faster training and inference. Implementation is done in PyTorch~\cite{paszke_pytorch_2019}, and training is performed using mean squared error (MSE) loss:

\begin{equation}
\text{MSE}_{\text{Loss}} = \frac{1}{n} \sum_{i=1}^{n} (y_i - \hat{y}_i)^2
\end{equation}

\noindent The training dataset consists of segmented, preprocessed images (see Section~\ref{subsec:DataPP}), each labeled with a corresponding set of control parameters (e.g. injection angle, mass flow rate). Rather than training on full images, we decompose each image into pixel-level entries (see Figure~\ref{PixCOIN_Method}, \textit{Training}), with each training sample comprising: (i) normalized pixel coordinates $(x, y)$, (ii) associated physical parameters $\mathbf{p}$, and (iii) the corresponding target pixel intensity. These entries are randomly sampled during training to ensure broad exposure across the dataset and to prevent overfitting to image-level structure.

\subsection{Inference and Image Reconstruction}
At inference time, the physical input parameters $\mathbf{p}$ are fixed to a desired operating point. The network sequentially evaluates all spatial coordinates $(x, y)$ in the image domain, producing corresponding intensity predictions (see Figure~\ref{PixCOIN_Method}, \textit{Inference}). These scalar outputs are then assembled into a 2D image representing the predicted film distribution.

This process effectively constructs a continuous, conditionally generated map of pixel intensities - interpretable as a probabilistic estimate of film density or presence at each location - without requiring spatial smoothing, continuity constraints, or handcrafted geometric features. By leveraging both physical and spatial inputs, PixCOIN offers a highly flexible and efficient surrogate model for experimental visualization, enabling the interpolation of unseen operating points with minimal physical testing.

\begin{figure}[b]
\centering
\subfigure[]{ 
  \makebox[0.65\textwidth]{ 
    \includegraphics[height=45mm]{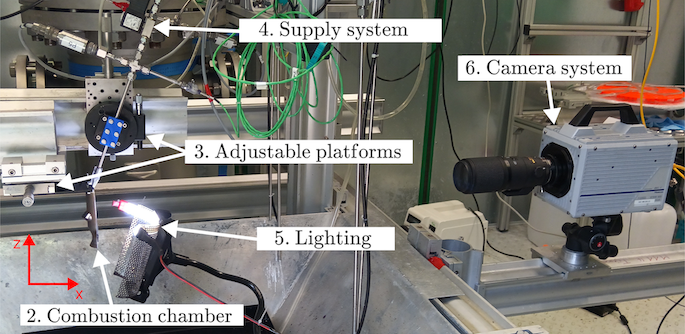}
  }
  \label{figSETUP:Versuchsaufbau_1}
}
\hspace{-15mm}
\subfigure[]{ 
  \makebox[0.3\textwidth]{ 
    \includegraphics[height=45mm]{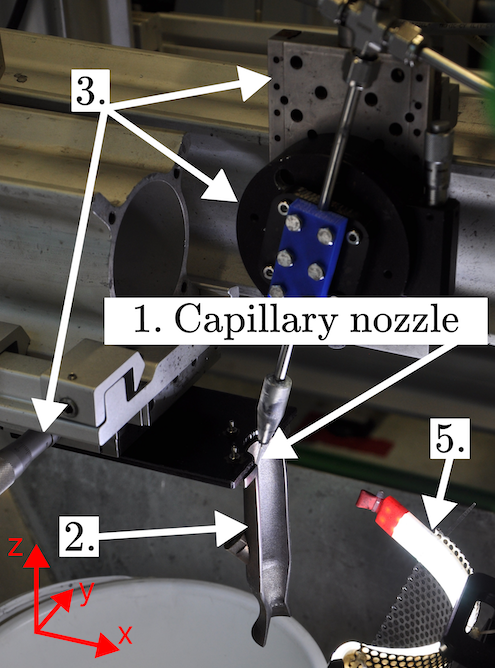}
  }
  \label{figSETUP:Versuchsaufbau_2}
}
\caption{Experimental setup for conducting cold flow tests of a film cooling setup with a single coolant injector on a semi-opened combustion chamber. (a) shows the full setup whilst (b) provides a close-up view of the combustion chamber, injector and lighting setup.}
\label{figSETUP:main}
\end{figure}

\section{Application: Film Cooling Investigations}

To demonstrate the practical utility of the proposed machine learning method, we apply it to a real-world film cooling study in an experimental propulsion context. The goal is to interpolate fluid film behavior across varying operating conditions using a significantly reduced number of physical tests. This section describes the experimental test setup and the associated data preparation procedures.

\subsection{System Setup}\label{sec:SysSetup}
A modular experimental setup was developed to investigate fluid film behavior in a simplified film cooling scenario. These initial studies focus on single-injector configurations, serving as a foundational case for later investigations involving multiple injectors and increased geometric complexity to better approximate conditions in fully integrated propulsion systems.

The test setup consists of a semi-open combustion chamber, providing direct optical access to the film layer. Fluid is injected through a capillary nozzle mounted on a mechanically adjustable platform. The injector assembly allows for variations in injection angle, axial position, and lateral offset, thereby enabling the exploration of multiple geometric degrees of freedom. With the supply system setup, we were able to control the mass flow for different injectors through the set pressure. Figure~\ref{figSETUP:main} provides an overview of the setup design.

High-speed imaging was conducted using a Photron FASTCAM SA1.1 camera equipped with a Nikon ED AF MICRO NIKKOR lens. The sensor captured grayscale images at \texttt{1024 x 1024} resolution and 50 FPS. Exposure time was varied between 1/50 and 1/300 seconds depending on lighting conditions and chamber surface reflectivity. To enhance image contrast, we employed uniform directional lighting as illustrated in Figure~\ref{figSETUP:Versuchsaufbau_2}.

As a surrogate fluid, demineralized water was selected for its ease of handling, replacing high-test peroxide (HTP) which is hazardous and difficult to manage in a lab setting. To increase optical contrast, Pelikan 4001 ink was added at a 100:4 water-to-ink volume ratio. Table~\ref{tab:rheolog} summarizes the rheological properties of the fluids, including density, dynamic viscosity, and interfacial tension. These properties were measured at 25\,\textdegree C using calibrated lab equipment (Mettler-Toledo Easy D40, Anton Paar LOVIS 2000 M, and Kr\"uss EASY DYNE). Notably, the addition of ink significantly reduced the interfacial tension, further differentiating the surrogate fluid from actual HTP. While this discrepancy may influence film dynamics, our primary focus is on demonstrating the model's interpolation capabilities; future work will consider more representative fluids.

\begin{table}[tb]
\centering
\caption{Comparison of various rheological properties of HTP, demineralized (DI) water, and ink solution in demineralized water. Reference values for HTP density and dynamic viscosity are taken from Teuffel et al.~\cite{teuffel_HTP_2024} for 97~wt.\% at 24.85 \textdegree C; interfacial tension is from Phibbs et al.~\cite{phibbs_hydrogen_1951} for 97.86~wt.\% at 20 \textdegree C.}
\label{tab:rheolog}
\medskip
\begin{tabular}{@{}llccc@{}}
\toprule
\textbf{Property} & \textbf{Unit} & \textbf{HTP} & \textbf{DI Water} & \textbf{Ink Solution} \\
\midrule
Density & \si{g.cm^{-3}} & 1.4340 & 0.9969 & 0.9975 \\
Dynamic viscosity & \si{mPa.s} & 1.080 & 0.882 & 0.888 \\
Interfacial tension & \si{mN/m} & 79.88 & 71.05 & 62.10 \\
\bottomrule
\end{tabular}
\end{table}

\begin{figure}[tb]
\centering
\subfigure[]{
  \makebox[0.2\textwidth]{ 
    \includegraphics[height=23.28 mm]{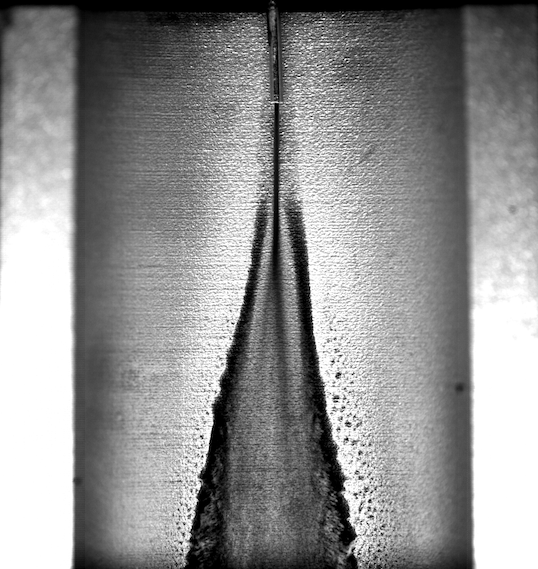}
  }
  \label{subsetup:1}
}
\hspace{-5mm}
\subfigure[]{
  \includegraphics[width=0.198\linewidth]{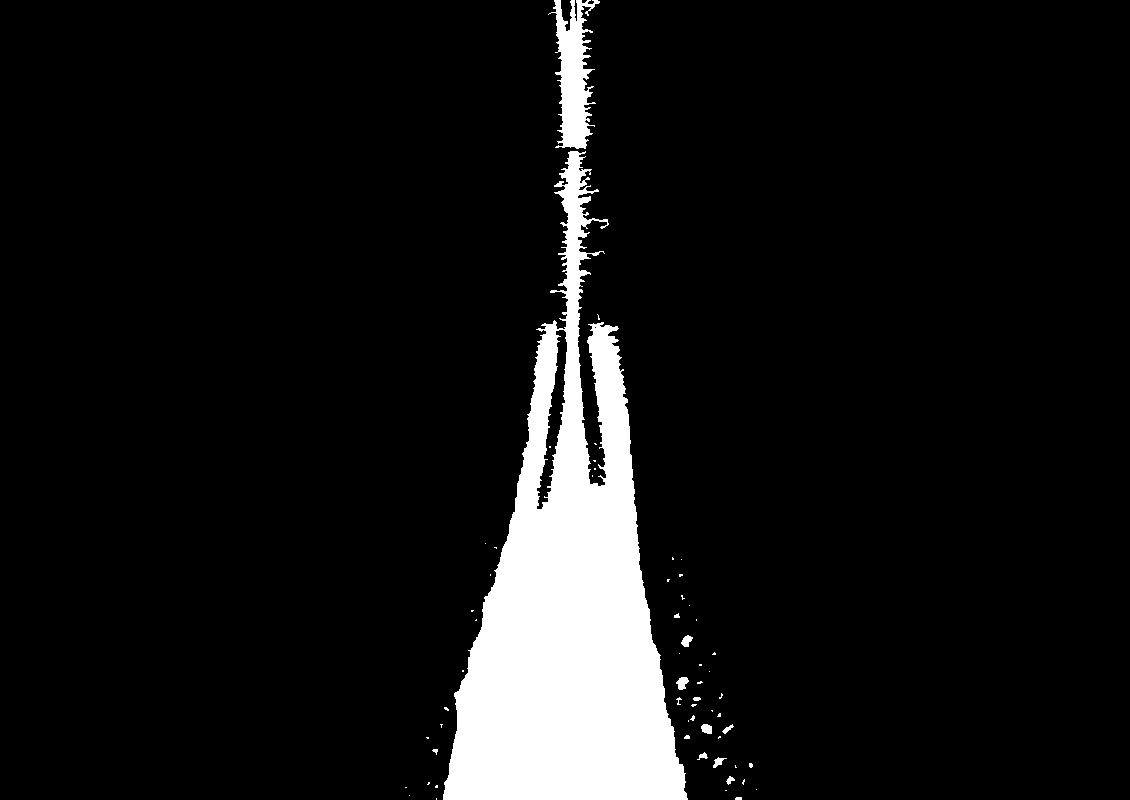}
  \label{subsetup:2}
}
\subfigure[]{
  \includegraphics[width=0.2\linewidth]{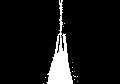}
  \label{subsetup:3}
}
\subfigure[]{
  \includegraphics[width=0.2\linewidth]{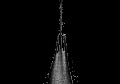}
  \label{subsetup:4}
}
\caption{Different image processing steps for an example image of a coolant film with our surrogate ink and demineralized water solution. (a) shows the image captured by the imaging system, (b) the binarization result, (c) the downsampled binary image and (d) a grayscale image, obtained by using (c) to mask (a).}
\label{subsetup:gesamt}
\end{figure}

\subsection{Data Preparation and Preprocessing}\label{subsec:DataPP}
The raw dataset consists of 4800 grayscale images capturing the fluid film behavior across all experimental conditions. To ensure consistent and accurate learning signals for the interpolation model, we implemented a structured preprocessing pipeline comprising geometric normalization, temporal denoising, segmentation, and resolution adjustment.

\vspace{1em}
\noindent \textbf{Geometric Normalization.} Images were first cropped to a fixed height and to the full width of the combustion chamber section used in each test. Due to the semi-cylindrical geometry of the opened combustion chamber, width variations introduce nonlinear warping. To account for this, we perform geometric rectification based on the method of Zankevich~\cite{zankevich_unwrap_labels}. The upper and lower boundaries of the chamber are modeled as elliptical arcs, parameterized by \(x = a \cos\phi\), \(y = b \sin\phi\), and then rotated into image space. A regular grid of source coordinates is interpolated across this shape, and each quadrilateral cell is transformed using a local homography via OpenCV's~\cite{opencv_library} \texttt{cv2.getPerspectiveTransform}.

\vspace{1em}
\noindent \textbf{Temporal Denoising.} To mitigate transient noise and lighting fluctuations, we compute a temporal mean over a stack of 10 consecutive images per operating point. This averaging interval was selected to balance noise reduction with the preservation of short-timescale features such as droplet formation. Longer intervals would risk blurring out relevant dynamic behavior of the fluid film.

\vspace{1em}
\noindent \textbf{Segmentation.} The fluid film is isolated using a two-step image processing approach. First, a background subtraction is applied using the \texttt{BackgroundSubtractorMOG2} method in OpenCV, referencing a background image taken under static conditions. This method is sensitive to subtle lighting variations caused by the absorptive properties of the ink-dyed water, which introduces artifacts. To minimize false positives, conservative parameter settings were used, which reduced noise at the cost of incomplete segmentation. 

Second, we apply adaptive thresholding to produce a binary image, followed by contour analysis to retain only the largest connected region, assumed to represent the fluid film. The final segmentation is produced by combining both methods using a logical OR operation, followed by minimal artifact filtering. An example of this segmentation can be seen in Figure~\ref{subsetup:2} for the fluid in Figure~\ref{subsetup:1}. While effective for this study, future work may benefit from a machine learning-based segmentation approach for improved robustness.

\vspace{1em}
\noindent \textbf{Resolution Normalization.} To ensure all input images are of uniform dimensions, images from tests involving narrower combustion chambers are symmetrically padded in width with background pixels. Given that our interpolation model operates on individual pixels, image resolution directly affects training and inference runtime. For this study, we downsample all images to \texttt{120 x 84} pixels (see Figure~\ref{subsetup:3}), balancing computational efficiency with the preservation of major structural features such as film contours and large droplets. Higher resolutions can be retained in practical deployments where finer spatial detail is required.

\section{Experimental Studies}
In this chapter, we evaluate the performance of the proposed interpolation method using both synthetic and real-world datasets derived from film cooling experiments. Our primary goal is to assess the model's ability to interpolate high-dimensional parameter spaces from sparse experimental data.

To ensure a focused and interpretable evaluation, we varied five representative operating parameters (see Table~\ref{tab:filmcooling_parameters}) that capture the dominant physical effects influencing film formation. This selection provides sufficient complexity to challenge the model's generalization capabilities while keeping the parameter space tractable for systematic investigation.

As the datasets consist of binary images, the model predicts scalar pixel intensities within the range \([0, 255]\). We considered applying a sigmoid activation function in the model's output layer to enforce this range implicitly. However, experiments showed no performance benefit from sigmoid activation compared to simply clipping output values outside the valid range. Consequently, we omitted the sigmoid to maintain a simpler architecture and avoid the risk of vanishing gradients.

\begin{table}[b]
\centering
\caption{List of the investigated parameters in the cold flow tests with all settings that were used during testing.}
\label{tab:filmcooling_parameters}
\medskip
\begin{tabular}{@{}lll@{}}
\toprule
\textbf{Parameter} & \textbf{Unit} & \textbf{Value} \\
\midrule
Manufacturing method        & --    & conventional; 3D-printed \\
Combustion chamber diameter & mm    & 20; 30; 45; flat surface \\
Injector nozzle diameter    & mm    & 0.3; 0.5 \\
Injection angle $\varphi$  & \textdegree     & 0; 5; 10; 15; 30; 45 \\
Pressure                    & bar   & 5; 10; 15; 20; 25 \\
\bottomrule
\end{tabular}
\end{table}

\subsection{Datasets}
\textbf{Synthetic Data.} To create a controlled benchmark for model evaluation, we generated synthetic images using the parametric function

\begin{equation}
f(x) = u_1 \cdot e^{- \left( u_2 \cdot (x - u_3) \right)^2} \cdot \cos(u_4 \cdot x) + u_5,
\label{eq_synthImages}
\end{equation}

\noindent with parameters \( u_1 \) through \( u_5 \). We constrained the input domain to \( x \in [-0.4, 0.6] \) and generated three types of images: (1.) a binary plot of the function; (2.) a filled-in integral representation; and (3.) a grayscale gradient version of the integral. All images were sized at \texttt{50 x 50} pixels. Doubling the resolution to \texttt{100 x 100} yielded no notable performance improvement, so we retained the smaller resolution for computational efficiency.

To simulate realistic uncertainty, random noise of \( \pm 5\% \) was added to the parameter values \emph{after} the image was generated, simulating noisy metadata. Although five setpoints were available for each parameter, only \( u_1 \), \( u_2 \), and \( u_3 \) were varied in early evaluations, using the ranges \( u_1 \in [1.5, 8.5] \), \( u_2 \in [1.7, 3.1] \), and \( u_3 \in [1.0, 1.3] \), while fixing \( u_4 = 0.9 \) and \( u_5 = 0.0 \). Furthermore, to evaluate the effect of feature scaling on model performance, the parameter ranges were mapped to different orders of magnitude for normalization testing.

\vspace{1em}
\noindent \textbf{Real Data.} The real-world dataset was obtained from the film cooling test setup described in Section~\ref{sec:SysSetup}, with images processed as outlined in Section~\ref{subsec:DataPP}. Each image was downsampled to a resolution of \texttt{120 x 84} pixels. We investigated the five parameters listed in Table~\ref{tab:filmcooling_parameters}, using all possible parameter combinations, resulting in 480 unique operating points. Ten individual images were recorded for each operating point, resulting in a total of 4800 data points.

The dataset was split into training, validation, and test subsets using a 70:20:10 ratio. Additionally, to validate the feasibility of using PixCOIN on continuous-valued grayscale data, a separate grayscale dataset with fewer parameter variations was generated and evaluated.

\subsection{Metrics}
To evaluate the performance of the proposed model, we consider both qualitative visual inspection and a set of quantitative image similarity metrics. In generative image synthesis tasks, commonly adopted metrics include the Peak Signal-to-Noise Ratio (PSNR) and the Structural Similarity Index Measure (SSIM)~\cite{hore_image_2010, sara_image_2019}. In addition to these, we report Root Mean Squared Error (RMSE), Cosine Similarity, and Perceptual Hashing, providing a comprehensive assessment from both pixel-wise and perceptual perspectives.

\vspace{1em}
\noindent \textbf{Root Mean Squared Error (RMSE).} RMSE~\cite{sara_image_2019} provides a pixel-wise deviation measure between the predicted image \( Y \) and the ground truth image \( X \), normalized by the pixel value range:

\begin{equation}
\text{RMSE} = \sqrt{ \frac{1}{NM} \sum_{i=1}^{N} \sum_{j=1}^{M} \left( \frac{Y_{i,j} - X_{i,j}}{255} \right)^2 },
\label{rmse_eq}
\end{equation}

\noindent where \( N \times M \) denotes the image resolution. Lower RMSE values indicate higher similarity.

\vspace{1em}
\noindent \textbf{Peak Signal-to-Noise Ratio (PSNR).} PSNR~\cite{deshpande_video_PSNR_2018} expresses the ratio between the maximum possible pixel value and the mean squared error (MSE), serving as a logarithmic measure of reconstruction quality:

\begin{equation}
\text{PSNR} = 10 \cdot \log_{10} \left( \frac{\text{MAX}^2}{\text{MSE}} \right),
\end{equation}

\noindent with \(\text{MAX} = 255\) for 8-bit images. Higher PSNR values reflect better similarity, although it remains sensitive to small pixel-level variations.

\vspace{1em}
\noindent \textbf{Structural Similarity Index Measure (SSIM).} SSIM~\cite{hore_image_2010} quantifies perceptual similarity by comparing local patterns of luminance, contrast, and structure:

\begin{equation}
\text{SSIM} = \frac{(2\mu_X\mu_Y + c_1)(2\sigma_{XY} + c_2)}{(\mu_X^2 + \mu_Y^2 + c_1)(\sigma_X^2 + \sigma_Y^2 + c_2)},
\end{equation}

\noindent where \( \mu \), \( \sigma \), and \( \sigma_{XY} \) denote the mean, standard deviation, and covariance of the images, respectively. Constants \( c_1 \) and \( c_2 \) are used to stabilize the computation. SSIM values range from 0 to 1, with values closer to 1 indicating greater similarity.

\vspace{1em}
\noindent \textbf{Cosine Similarity.} Cosine similarity~\cite{lahitani_cosine_2016} measures the angular similarity between image vectors reshaped into one-dimensional arrays. Values closer to 1 indicate greater directional alignment and therefore higher similarity.

\vspace{1em}
\noindent \textbf{Perceptual Hashing.} Perceptual hashing~\cite{farid_overviewPercHash_2021} transforms an image into a hash value such that visually similar images yield similar hashes. The similarity is computed by the Hamming distance between these hash values, providing a compact perceptual descriptor.

\vspace{1em}
\noindent \textbf{Implementation Details.} SSIM and PSNR were computed using the \texttt{scikit-image} library~\cite{walt_scikit-image_2014}. Cosine similarity is implemented following the method of Okonji~\cite{okonji_cosine_similarity}, while perceptual hashing is based on the library by Buchner~\cite{buchner_imagehash}.

\begin{figure}[tb]
\centering
\includegraphics[width=\textwidth]{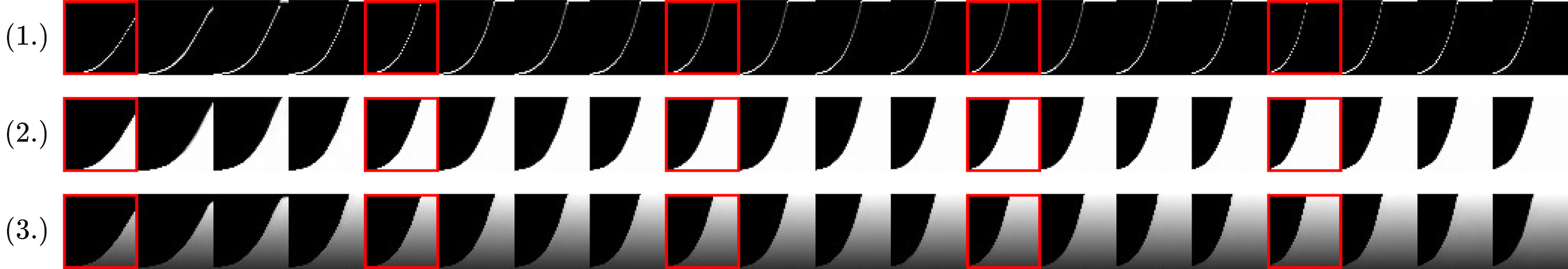}
\caption{PixCOIN regression results on synthetic images for datasets of (1.) a binary function plot, (2.) a representation with a filled-in integral and (3.) a version of this integral with a grayscale gradient. All red-framed images are training images, whilst all other images were generated using PixCOIN.}
\label{synth_interpolation}
\end{figure}

\begin{table}[b]
\centering
\caption{Comparison of RMSE values of models trained on different datasets to analyze the difference between interpolation and extrapolation capabilities of PixCOIN.}
\label{tab:rmse_interp_extrap}
\medskip
\begin{tabular}{@{}lccc@{}}
\toprule
\textbf{Levels of Parameter $u_3$} & \textbf{Reference} & \textbf{Interpolation} & \textbf{Extrapolation} \\
\midrule
Level 1 (1.00) & 6.88 & 6.41 & \textbf{17.26} \\
Level 2 (1.15) & 5.80 & \textbf{7.59} & 5.40 \\
Level 3 (1.30) & 4.13 & 3.80 & 4.16 \\
\bottomrule
\end{tabular}
\end{table}

\begin{figure}[tb]
\centering
\includegraphics[width=\textwidth]{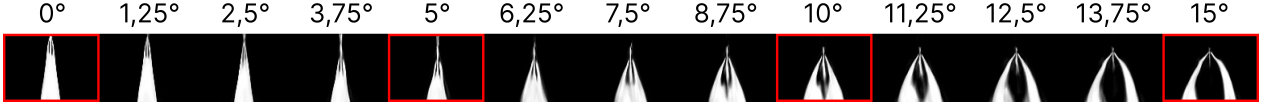}
\caption{PixCOIN interpolation results on real images. This one-dimensional depiction shows varying the injection angle $\varphi$. All red-framed images are training images, whilst all other images were generated using PixCOIN.}
\label{real_interpolation}
\end{figure}

\begin{table}[tb]
\centering
\caption{Quantitative evaluation results for different reductions of the number of parameter settings of the parameter mass flow. Results for the model trained on the full dataset (mass flow all 5 possible settings) are included as a baseline.}
\label{tab:massflow_metrics}
\medskip
\begin{tabular}{@{}lccc@{}}
\toprule
\textbf{Metric} & \textbf{Baseline} & \textbf{3 Settings} & \textbf{2 Settings} \\
\midrule
RMSE & 0.069 $\pm$ 0.028 & 0.078 $\pm$ 0.031 & 0.090 $\pm$ 0.040 \\
SSIM & 0.942 $\pm$ 0.043 & 0.927 $\pm$ 0.051 & 0.909 $\pm$ 0.061 \\
PSNR & 23.871 $\pm$ 3.245 & 22.760 $\pm$ 3.313 & 21.727 $\pm$ 3.895 \\
Cosine Similarity & 0.575 $\pm$ 0.078 & 0.553 $\pm$ 0.082 & 0.555 $\pm$ 0.084 \\
Perceptual Hashing & 5.926 $\pm$ 3.951 & 5.814 $\pm$ 3.739 & 7.501 $\pm$ 3.778 \\
\bottomrule
\end{tabular}
\end{table}

\begin{table}[tb]
\centering
\caption{Quantitative evaluation results for further configurations with reduced datasets. Reduced number of parameter settings for (col. 1) the injection angle from 6 to 3 and (col. 2) all parameters to a maximum of 3 settings. Additionally (col. 3) results of a model trained on a randomly reduced dataset by 30\% of operating points.}
\label{tab:influence_factors_metrics}
\medskip
\begin{tabular}{@{}lccc@{}}
\toprule
\textbf{Metric} & \textbf{Angle 3 Set.} & \textbf{All Factors max. 3 Set.} & \textbf{30\% rand. Reduced} \\
\midrule
RMSE & 0.088 $\pm$ 0.049 & 0.116 $\pm$ 0.068 & 0.078 $\pm$ 0.036 \\
SSIM & 0.872 $\pm$ 0.069 & 0.892 $\pm$ 0.076 & 0.934 $\pm$ 0.049 \\
PSNR & 22.307 $\pm$ 4.411 & 20.135 $\pm$ 4.941 & 22.917 $\pm$ 3.652 \\
Cosine Similarity & 0.545 $\pm$ 0.091 & 0.543 $\pm$ 0.099 & 0.558 $\pm$ 0.081 \\
Perceptual Hashing & 7.930 $\pm$ 3.808 & 8.025 $\pm$ 4.009 & 6.122 $\pm$ 3.900 \\
\bottomrule
\end{tabular}
\end{table}

\begin{figure}[tb]
\centering
\includegraphics[width=\textwidth]{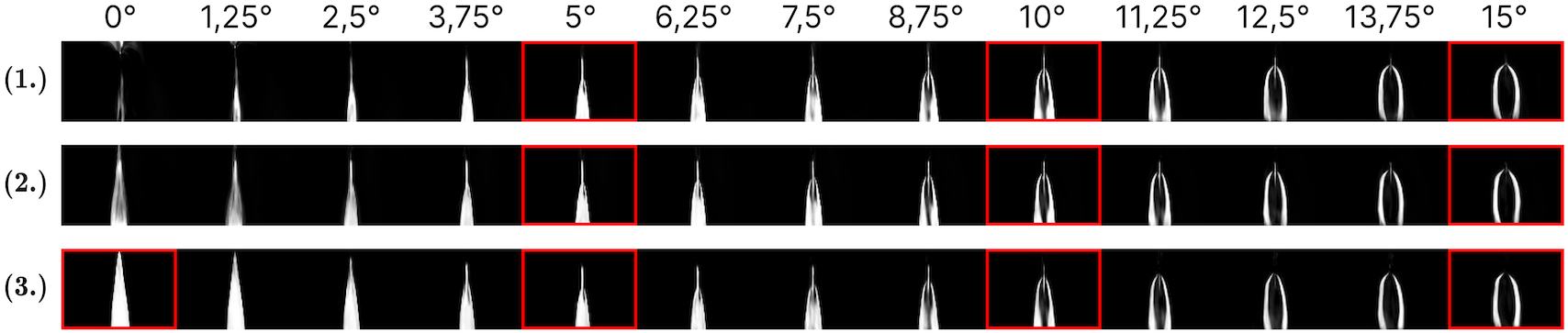}
\caption{PixCOIN regression results on real images showing the variation of the injection angle $\varphi$. Line (1.) shows extrapolation, where no operating point with $\varphi$ of 0 was given as training data. Line (2.) shows extrapolation where one single operating point with $\varphi$ of 0 was given as training data. All red-framed images are training images, whilst all other images were generated using PixCOIN.}
\label{interpolation_realData}
\end{figure}

\begin{figure}[tb]
\centering
\subfigure[]{
  \includegraphics[width=0.12\textwidth]{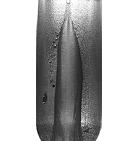}
  \label{sub12:1}
}
\subfigure[]{
  \includegraphics[width=0.12\textwidth]{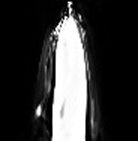}
  \label{sub12:2}
}
\subfigure[]{
  \includegraphics[width=0.12\textwidth]{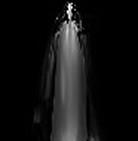}
  \label{sub12:3}
}
\subfigure[]{
  \includegraphics[width=0.12\textwidth]{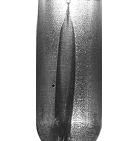}
  \label{sub12:4}
}
\subfigure[]{
  \includegraphics[width=0.12\textwidth]{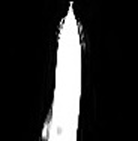}
  \label{sub12:5}
}
\subfigure[]{
  \includegraphics[width=0.12\textwidth]{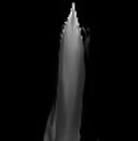}
  \label{sub12:6}
}
\caption{Real images that were cropped ((a) and (d)), generated by PixCOIN trained on binary data ((b) and (e)) and generated by PixCOIN trained on grayscale data ((c) and (f)). Different systematic errors are shown with (left) a drop that did not drain of during data sampling and (right) skewed flow due to a manufacturing defect in the injector capillary.}
\label{subfig12:gesamt}
\end{figure}

\subsection{Results}
Model training was conducted for both synthetic and real datasets over 50 epochs using a batch size of 64, a learning rate of 0.0001, and the Adam optimizer. An initial ablation study was performed on synthetic data to validate the core capabilities of the proposed PixCOIN architecture.

Figure~\ref{synth_interpolation} illustrates interpolation and extrapolation results across three synthetic datasets, each varying a single parameter (in this case \( u_1 \) from equation~\ref{eq_synthImages}). Red-framed images indicate training samples, while all other images are generated by the model. These visualizations demonstrate accurate regression of the underlying function, including fine details such as single-pixel-wide structures (row (1.)). The model's ability to generate grayscale outputs with smooth gradients is further evidenced in row (3.), indicating successful learning of both structural and tonal features.

To quantify the interpolation and extrapolation capabilities, Table~\ref{tab:rmse_interp_extrap} presents RMSE values for parameter \( u_3 \) under three evaluation configurations. Interpolation results in only a marginal increase in error, whereas extrapolation yields a significantly higher RMSE, consistent with expected limitations of data-driven extrapolation.

Encouraged by these results, preliminary experiments were conducted on real experimental data. We observed improvements in all quantitative metrics when both the input parameters and pixel coordinates were normalized. Additionally, performance remained stable when reducing the number of samples per operating point from 10 to 5, effectively halving the dataset size without loss in quality.


These findings informed the final model configuration described in Section~\ref{subsec:ArchitectureAndTraining}, using the identified hyperparameters and a reduced dataset of five samples per operating point. An exemplary interpolation of real test data over one parameter is shown in Figure~\ref{real_interpolation}.


\vspace{1em}
\noindent \textbf{Reduced experimental testing.}  
To evaluate the impact of dataset sparsity, we conducted experiments by reducing the number of parameter settings used during training. We investigate mass flow, being an indirect parameter dependent on both injection pressure and injector nozzle diameter, as shown in Table~\ref{tab:filmcooling_parameters}. The full dataset includes five discrete mass flow settings. We tested two reduced configurations: (i) three settings (minimum, maximum, and median), and (ii) two settings (minimum and maximum).

The results, summarized in Table~\ref{tab:massflow_metrics}, show that dataset reduction leads to a moderate degradation in image generation quality. Specifically, reducing to three settings resulted in a decrease in RMSE of only 0.95~\% and in SSIM of 1.54~\%, while the two-setting configuration caused RMSE and SSIM to drop by 2.16~\% and 3.28~\%, respectively. Despite these reductions, the performance remained within acceptable limits (errors below 10~\%), suggesting the model maintains good generalization even with sparse training data.

We further explored dataset reduction strategies in Table~\ref{tab:influence_factors_metrics}. In one test, only the injection angle $\varphi$ was limited to three settings, reducing the total number of ground truth operating points from 480 to 240. In another setup, all factors with more than three settings were truncated to three, leading to a reduced dataset of 162 tested operating points. Additionally, a randomly reduced dataset--removing 30~\% of operating points-was evaluated to compare systematic versus unsystematic reduction approaches.

\vspace{1em}
\noindent \textbf{Further Results.}  
The contrast between interpolation and extrapolation, first highlighted with synthetic data in Table~\ref{tab:rmse_interp_extrap}, is also evident in real data. Figure~\ref{interpolation_realData} illustrates this: in row (1.), the network is extrapolating an unseen injection angle (0\textdegree), resulting in noticeably lower image quality compared to row (3.), which contains interpolated cases. When one operating point with a similar angle was included during training (row (2.)), a qualitative improvement was observed. However, the corresponding quantitative gain over the full parameter space was minor, with RMSE improving by only 0.5~\% and SSIM by 0.02~\%.

Figure~\ref{subfig12:gesamt} illustrates transient behaviors, with haze-like patterns indicating overspray or droplet breakup-an emergent behavior interpreted as probabilistic estimates of local film presence. Moreover, Figure~\ref{sub12:2} reveals systematic biases embedded in the training data: a persistent fluid drop, not removed during consecutive testing runs, consistently reappears in generated images. Additional qualitative results are presented in Figures~\ref{sub12:3} and~\ref{sub12:6}, demonstrating the model's ability to capture real film cooling behavior using grayscale data. In particular, variations in grayscale intensities indicate film thickness variations.

In terms of computational efficiency, the selected medium-sized model required 7:02 minutes per epoch on average in training, measured on a NVIDIA RTX 3500 Ada Generation Laptop GPU. The larger model increased training time to 8:54 minutes per epoch. For the 30\%-reduced dataset, training time decreased to 4:52 minutes per epoch. Inference duration for a single image was 5.98 seconds with the medium model, and 8.77 seconds with the larger model, averaged over ten samples.

\section{Discussion}
We demonstrate overall strong performance of our proposed method in synthesizing high-fidelity images across a range of operating points. The network architecture exhibits robust generalization, achieving consistent results for both small (50 x 50 pixels for synthetic data) and moderately larger image sizes (120 x 84 pixels for real data). Nonetheless, for applications involving substantially larger image dimensions, architectural adaptations or scaling strategies may be necessary to maintain accuracy and efficiency.

\begin{figure}[b]
\centering
\includegraphics[width=12cm]{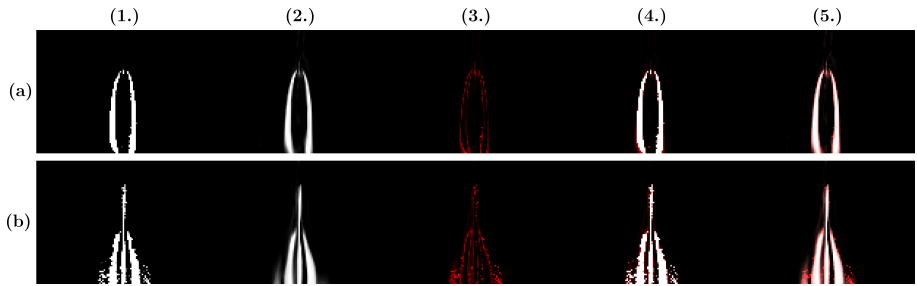}
\caption{Exemplary comparison of real (1.) and generated (2.) image data for two operating points. Column (3.) shows the difference of the foreground of images in (1.) and (2.) in red, with overlays of (3.) with (1.) shown in (4.) and (3.) and (2.) shown in (5.).}
\label{Bsp_ImgSimil}
\end{figure}

\begin{figure}[tb]
\centering
\includegraphics[height=65mm]{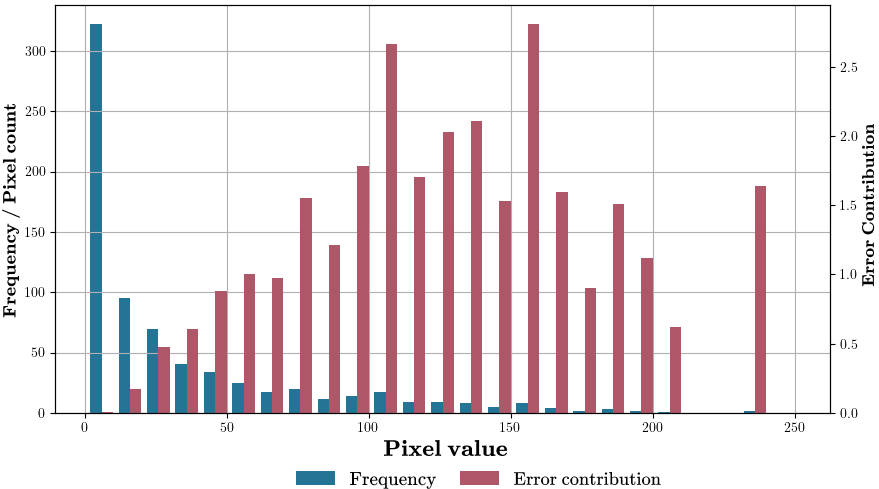}
\caption{Histogram of pixel values in bins of intensity values of multiples of 10 and their contribution to the RMSE-Value of the operating point shown in figure~\ref{Bsp_ImgSimil}(a)}
\label{plot_BP_1}
\end{figure}

\vspace{1em}
\noindent \textbf{Impact of Dataset Reduction.}
Performance degradation becomes apparent when reducing the number of measurements per operating point. As observed in Table~\ref{tab:massflow_metrics}, the reduction to two discrete mass flow settings results in a more significant loss in performance compared to a reduction to three settings. This observation suggests that the availability of a median reference point enables the model to learn a more nuanced, possibly nonlinear, representation of the physical process. Conversely, with only two reference values, the network might be forced into overly linear approximations of underlying relationships.

Despite this, the results show that the model retains impressive predictive accuracy even with substantially reduced datasets. For instance, when reducing the number of mass flow settings by 40\%, RMSE remains below 8\% and SSIM exceeds 92\%. When all parameters with more than three settings are capped at three, the dataset size is reduced by over two-thirds (66.25\%), yet RMSE remains under 12\% and SSIM above 89\%. A randomly reduced dataset (30\% fewer operating points) still results in RMSE below 8\% and SSIM above 93\%.

These findings indicate that the model captures the underlying physical relationships governing the film formation process effectively, allowing for meaningful reductions in experimental data requirements without major sacrifices in performance.

\vspace{1em}
\noindent \textbf{Composition of the RMSE.}
To provide further insight into the nature of the reported RMSE values in Tables~\ref{tab:massflow_metrics} and~\ref{tab:influence_factors_metrics}, we analyze two representative operating points shown in Figure~\ref{Bsp_ImgSimil}. Each row presents the reference (measured) image (col. (1.)), the corresponding synthesized image (col. (2.)), the absolute pixel-wise error (in red, col. (3.)), and overlayed visualizations highlighting discrepancies between the ground truth and synthesized data (cols. (4.) and (5.)).

Visual inspection reveals that the primary sources of error are located at the interfaces between foreground and background regions, as well as in dynamic or transient regions such as the droplet zone (see row (b)). Notably, these errors are localized and often limited to one-pixel-wide areas.

To quantify pixel-level error contributions, we compute the cumulative influence of different intensity levels on the overall RMSE. As shown in Figure~\ref{plot_BP_1}, the contribution of each intensity class $I_c$ is calculated as

\begin{equation}
\text{Error Contribution}_{I_c} = \text{Pixel Frequency} \cdot \left( \frac{\text{Class Intensity Level}}{255} \right)^2
\end{equation}

\noindent where the pixel frequency is given as the number of faulty pixels per bin. This formulation is derived from Equation~\ref{rmse_eq}, reflecting how each intensity class contributes to the overall squared error:

\begin{equation}
\sum_{i=1}^{n_{I_c}} \text{Error Contribution}_i \equiv \sum_{i=1}^{N} \sum_{j=1}^{M} (Y_{i,j} - X_{i,j})^2 \cdot \frac{1}{255}
\end{equation}

\noindent The histogram in Figure~\ref{plot_BP_1} confirms that the majority of pixels exhibit only minor intensity deviations, contributing little to the overall RMSE. The dominant error contribution stems from a small subset of pixels with large deviations. For example, the RMSE of 5.63\% for the case in Figure~\ref{Bsp_ImgSimil}(a) is largely attributed to such outliers. This analysis underscores that RMSE values of this magnitude often correspond to visually negligible discrepancies, thereby contextualizing the metrics reported in Tables~\ref{tab:massflow_metrics} and~\ref{tab:influence_factors_metrics}.

\vspace{1em}
\noindent \textbf{Challenges and Future Work.} While the PixCOIN approach demonstrates strong performance, several limitations and opportunities for improvement remain. Our study encountered systematic experimental inconsistencies in real-world test data, including manufacturing variations in capillary injectors and pre-wetting effects during sequential testing. These artifacts altered observed fluid film behavior and likely introduced bias into model training since they propagate into the ground truth data. Enhancing experimental controls--particularly for injector tolerances and test-sequence protocols--could mitigate such biases and improve model fidelity.


As discussed in Section~\ref{subsec:DataPP}, segmentation quality was limited, particularly around the RFZ. Segmentation errors led to occlusions in grayscale ground truth images, as we derived them using binary masks. Employing deep learning-based segmentation methods in future work may yield cleaner training data and more accurate grayscale synthesis.

While our grayscale image synthesis showed strong qualitative performance, future applications may benefit from generating full RGB outputs. This can be addressed by modifying the network's output layer to produce three-channel predictions.

Despite these challenges, the PixCOIN framework demonstrated robust performance even with reduced datasets, confirming its potential for efficient image regression under parametrized conditions. Further improvements in data quality and model architecture are expected to extend its applicability to broader use cases.

\begin{figure}[tb]
\centering
\includegraphics[width=\textwidth]{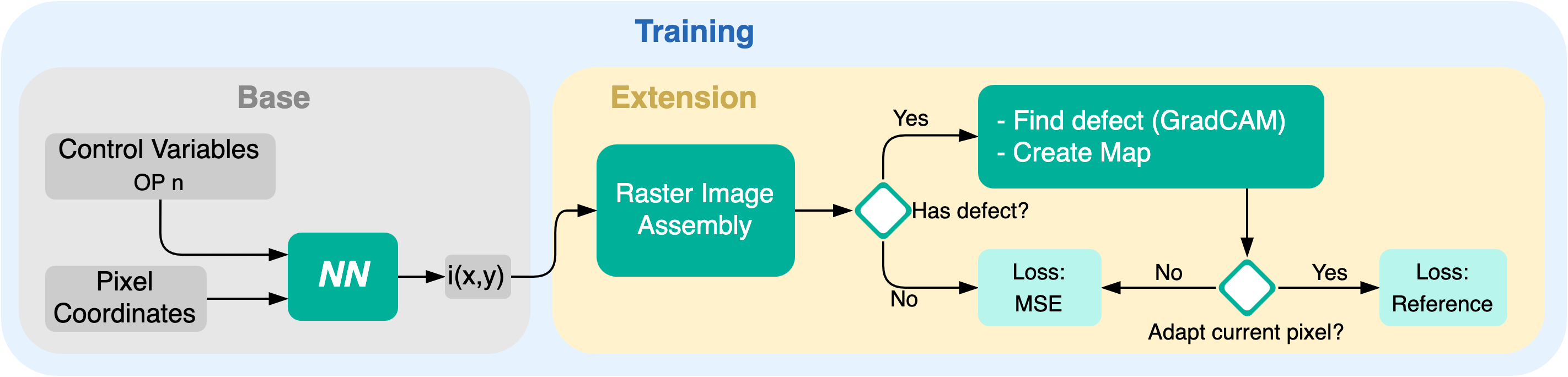}
\caption{Schematic flowchart of the extended training workflow to enable an expert informed adaptation.}
\label{Flussdiag_IML}
\end{figure}

\begin{figure}[tb]
\centering
\subfigure[]{
  \includegraphics[width=0.145\textwidth]{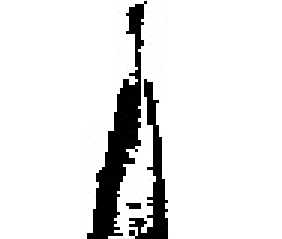}
  \label{sub111:1}
}
\subfigure[]{
  \includegraphics[width=0.145\textwidth]{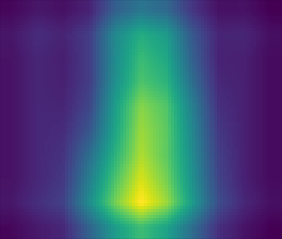}
  \label{sub111:2}
}
\subfigure[]{
  \includegraphics[width=0.145\textwidth]{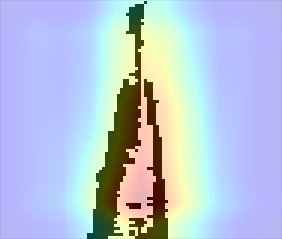}
  \label{sub111:3}
}
\subfigure[]{
  \includegraphics[width=0.145\textwidth]{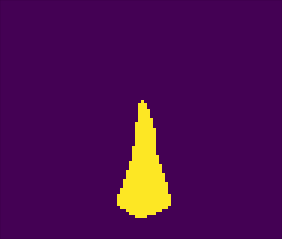}
  \label{sub111:4}
}
\subfigure[]{
  \includegraphics[width=0.145\textwidth]{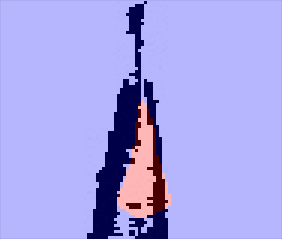}
  \label{sub111:5}
}
\caption{GradCAM activation (b) on an inverted image (a) from the classifier to detect zones contributing to a classification as a faulty segmentation. In order to create a discrete map the activation is binarized (d) and visualized in an overlay with (a) in (e).}
\label{subfig111:gesamt}
\end{figure}

\begin{figure}[tb]
\centering
\subfigure[]{
  \includegraphics[width=0.12\linewidth]{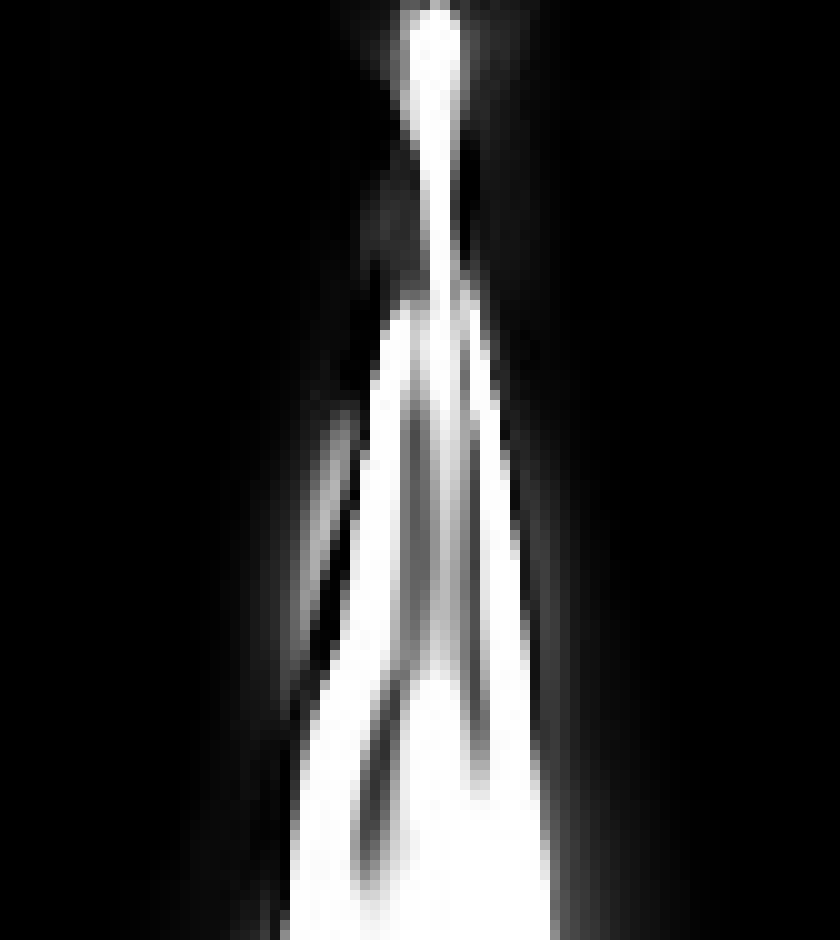}
  \label{sub6:1}
}
\subfigure[]{
  \includegraphics[width=0.12\linewidth]{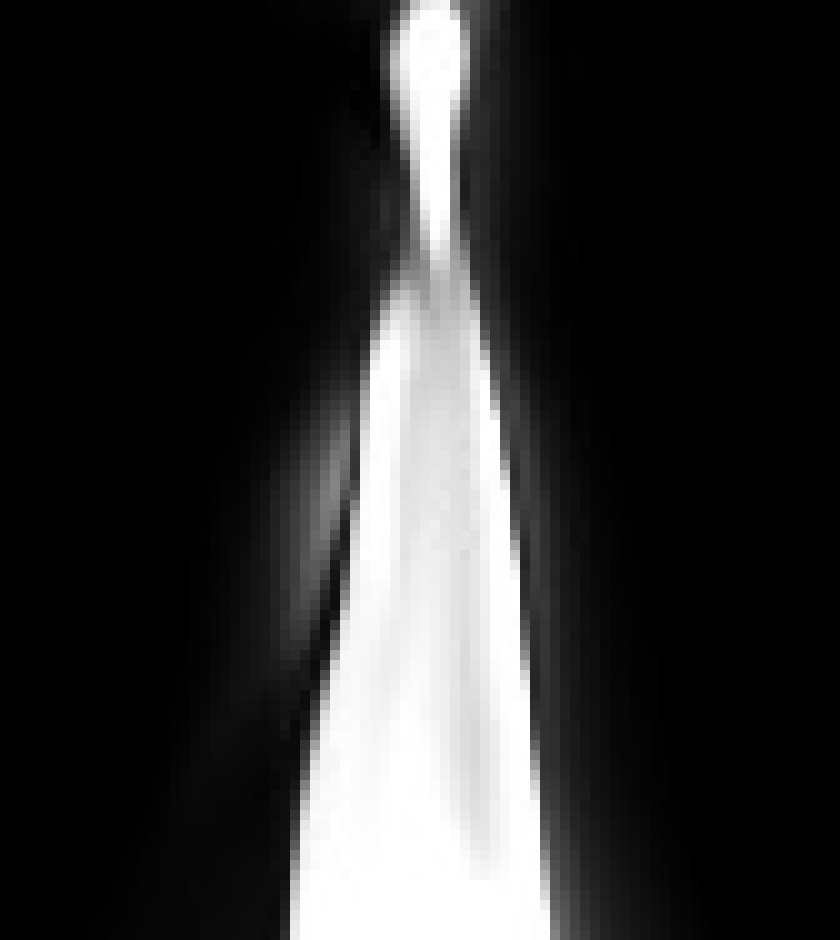}
  \label{sub6:2}
}
\subfigure[]{
  \includegraphics[width=0.12\linewidth]{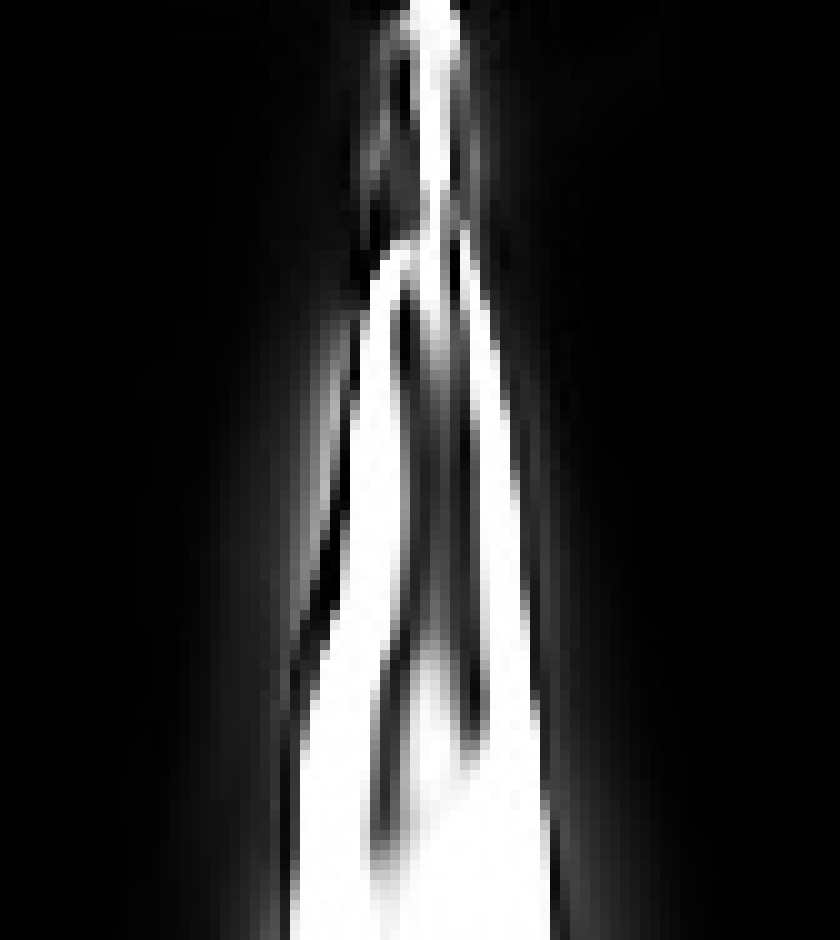}
  \label{sub6:3}
}
\subfigure[]{
  \includegraphics[width=0.12\linewidth]{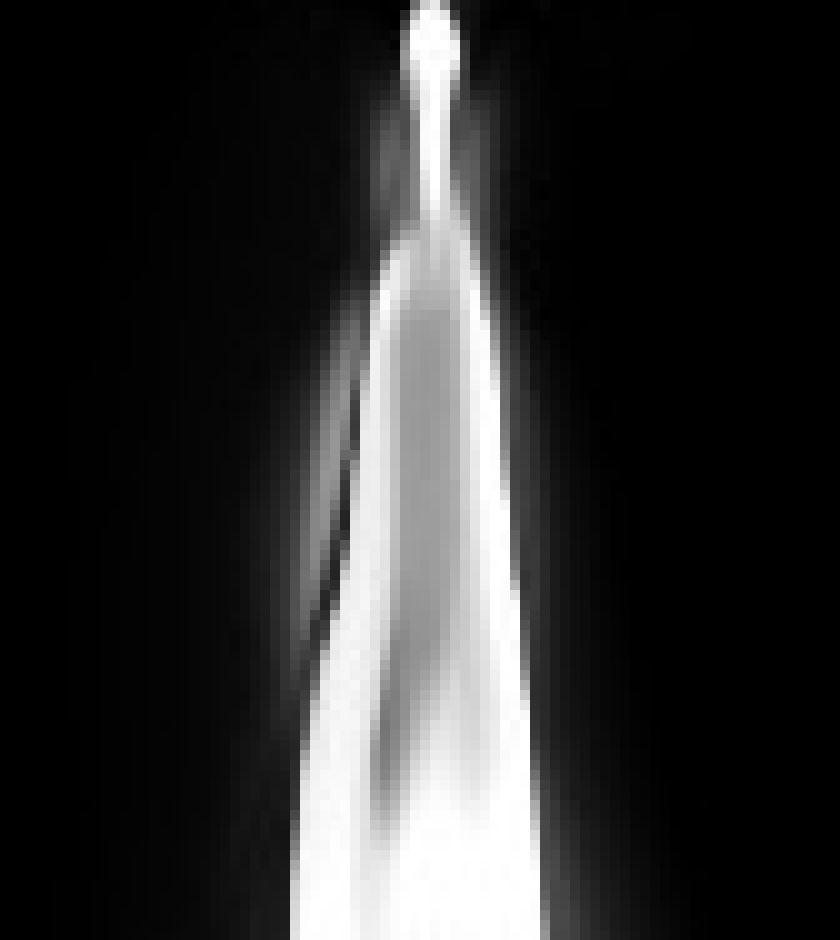}
  \label{sub6:4}
}
\caption{PixCOIN image generation results for two operating points without ((a), (c)) and with ((b), (d)) the IML extension for segmentation faults in the RFZ.}
\label{subfig6:gesamt}
\end{figure}

\section{Extension: Pixel-level Informed Adaptations}
Motivated by the segmentation challenges in the RFZ, we propose an informed machine learning (IML) extension to the PixCOIN framework that integrates expert knowledge to guide the image synthesis process. While demonstrated here for a specific use-case, the underlying concept is broadly applicable to other domains where targeted correction of image regions is desirable.

\vspace{1em}
\noindent \textbf{Concept.}
Expert knowledge was used to define regions (e.g., the RFZ) where fluid presence is expected but not accurately captured in the training data due to segmentation errors. During training, we detect such defects in synthesized images and selectively adapt the training strategy to correct them-without modifying the base architecture.

\vspace{1em}
\noindent \textbf{Procedure.}
As outlined in Figure~\ref{Flussdiag_IML}, the extension augments the standard FFNN by dynamically modifying the loss function based on pixel-level feedback. No changes to the network architecture are required.

To perform image-level reasoning, we first execute forward passes across all pixels of an image. Once a full image is synthesized, it is assessed for segmentation faults using a lightweight CNN classifier (4 convolutional layers + 2 fully connected layers, ReLU activations, and softmax output). The classifier is trained on a dataset of 250 images with faulty segmentations and 500 correctly segmented examples. To ensure that the classifier operates on sufficiently meaningful inputs, image-level assessment is deferred until after the fifth training epoch, once the generated images have reached a usable quality threshold.

If a defect is detected, we localize it using GradCAM~\cite{selvaraju_grad-cam_2017} on the final convolutional layer of the classifier. The resulting heatmap is binarized to produce a pixel-wise fault map (Figure~\ref{sub111:4}). During backpropagation, pixels flagged as faulty are trained using a modified loss function:

\begin{equation}
\text{Loss}_{\text{Ref}} = \frac{1}{n} \sum_{i=1}^{n} (y_i - r)^2
\end{equation}

\noindent where $r$ is a reference intensity value representing the expected foreground. All other pixels continue to use the standard MSE loss with ground truth values.

\vspace{1em}
\noindent \textbf{Results.}
Figure~\ref{subfig6:gesamt} shows two example outputs where the adaptation significantly improved synthesis in faulty RFZ regions. The extension successfully reclassified background artifacts into foreground, enhancing visual coherence. While the binary GradCAM maps are not perfect (Figure~\ref{sub111:5}), they provide sufficiently accurate guidance. For instance, in Figure~\ref{sub6:4}, partial detection in the lower-left RFZ limits correction-mainly due to a conservative threshold. Lowering the threshold could improve coverage but may distort the broader fluid film geometry.

Overall, this extension demonstrates that incorporating expert guidance at the pixel level through targeted loss modulation can significantly improve synthesis in critical regions. It offers a promising direction for future refinement and broader applicability in domain-informed image generation.

\section{Conclusion}
In the development of advanced propulsion systems, experimental validation remains a critical step to ensure the reliability and performance of combustion chamber components. However, such experiments are often costly and time-consuming. In this work, we introduced PixCOIN, a neural regression approach for synthesizing physically plausible images conditioned on operating parameters, with a particular focus on reducing the number of required experimental measurements.

Through extensive evaluation on both synthetic and real datasets, we demonstrated that PixCOIN can generate high-quality images even with significantly reduced training data. The method captures key physical behaviors and maintains strong performance across interpolation scenarios. We also introduced an extension that integrates expert knowledge for pixel-level adaptations, enabling targeted correction of faulty image regions due to segmentation errors.

Future work will explore more robust and automated segmentation strategies, including deep learning-based preprocessing. We also aim to expand the method to RGB image data and incorporate temporal dependencies to model transient behaviors. Further developments of the informed adaptation framework may include alternative guidance signals or multi-stage training strategies, enhancing its generalization and applicability across domains.

%

%
%
\bibliographystyle{plain}

\end{document}